\documentclass[10pt,twocolumn,letterpaper]{article}

\usepackage{cvpr}              

\usepackage{booktabs}
\usepackage{times}
\usepackage{epsfig}
\usepackage{graphicx}
\usepackage{amsmath}
\usepackage{amssymb}
\usepackage{algorithm}
\usepackage{algorithmic}
\usepackage{bm, multirow}
\usepackage[super]{nth}
\usepackage{enumitem}
\usepackage{color,soul}

\usepackage{tabularx} 
\usepackage{rotating}
\newcolumntype{L}[1]{>{\hsize=#1\hsize\raggedright\arraybackslash}X}%
\newcolumntype{R}[1]{>{\hsize=#1\hsize\raggedleft\arraybackslash}X}%
\newcolumntype{C}[1]{>{\hsize=#1\hsize\centering\arraybackslash}X}%

\DeclareMathOperator*{\argmin}{arg\,min}
\graphicspath{{figs/}}

\usepackage[pagebackref,breaklinks,colorlinks]{hyperref}

\usepackage[capitalize]{cleveref}
\crefname{section}{Sec.}{Secs.}
\Crefname{section}{Section}{Sections}
\Crefname{table}{Table}{Tables}
\crefname{table}{Tab.}{Tabs.}

\def\cvprPaperID{*****} 
\def\confName{CVPR}
\def\confYear{2022}

\begin{document}

\title{Update Compression for Deep Neural Networks on the Edge}

\author{Bo Chen$^1$ \quad Ali Bakhshi$^2$ \quad Gustavo Batista$^2$ \quad  Brian Ng$^1$ \quad Tat-Jun Chin$^1$\\
$^1$The University of Adelaide \quad $^2$University of New South Wales
}

\maketitle

\begin{abstract}
An increasing number of artificial intelligence (AI) applications involve the execution of deep neural networks (DNNs) on edge devices. Many practical reasons motivate the need to update the DNN model on the edge device post-deployment, such as refining the model, concept drift, or outright change in the learning task. In this paper, we consider the scenario where retraining can be done on the server side based on a copy of the DNN model, with only the necessary data transmitted to the edge to update the deployed model. However, due to bandwidth constraints, we want to minimise the transmission required to achieve the update. We develop a simple approach based on matrix factorisation to compress the model update---this differs from compressing the model itself. The key idea is to preserve existing knowledge in the current model and optimise only small additional parameters for the update which can be used to reconstitute the model on the edge. We compared our method to similar techniques used in federated learning; our method usually requires less than half of the update size of existing methods to achieve the same accuracy. 
\end{abstract}

\section{Introduction}
\label{sec:intro}

The significant progress in AI has been due in large part to the resurgence of deep neural networks (DNNs), particularly convolutional neural networks (CNNs) and variants thereof. As the complexity and breadth of problems that are solvable in an end-to-end manner by DNNs increase, the DNN architectures (``models'') themselves are becoming deeper~\cite{simonyan2015very,huang2017densely,he2016identity,he2016deep} and wider~\cite{szegedy2015going,Zagoruyko2016wide,szegedy2016rethinking}, which demand greater computational resources to execute.

On the other hand, there is a trend to deploy DNNs in the field through edge computing devices, such as embedded GPUs and FPGAs. This is due to the growing popularity of concepts such as AI on the edge~\cite{luo2020edgenas,chen2019deep,li2018learning, ning2020mobile}, which emphasises processing closer to the source of the data to minimise latency and reduce data transfer, and embodied AI~\cite{gobieski2019intelligence,ramakrishnan2021exploration,chaplot2020object}, which focusses on building intelligent agents that can explore and interact with the environment.

The gap in computational capability between centralised processing systems and edge computing devices has motivated research in making complex DNNs feasible on the latter platforms. Significant attention has been devoted to compressing DNNs to reduce their memory consumption and/or accelerate their execution.  Major categories of such methods include pruning and sparsification~\cite{tung2018clip,zhuang2018discrimination,he2018amc,he2017channel,liu2017sphereface}, quantisation~\cite{nagel2020up,li2017training,alizadeh2020gradient,martinez2020training,10.1007/978-3-319-46493-0_32}, knowledge distillation~\cite{tang2019learning,polino2018model,xu2017training,chen2016net2net,hinton2015distilling}, low-rank optimisation~\cite{kim2016compression,garipov2016ultimate,peng2018extreme,chen2018adaptive,kim2019efficient,kossaifi2019t}, and architecture search~\cite{xie2018interleaved,sun2018igcv3,howard2019searching,ma2018shufflenet,zhang2018shufflenet,gordon2018morphnet,tan2019mnasnet,tan2019efficientnet,sandler2018mobilenetv2}. A common observation is that many complex DNNs are over-parametrised, hence allowing substantial compression~\cite{kahatapitiya2021exploiting, shang2016understanding, chen2019drop, qiu2021slimconv}.

\begin{figure}[t]
    \centering
    \includegraphics[width=\linewidth]{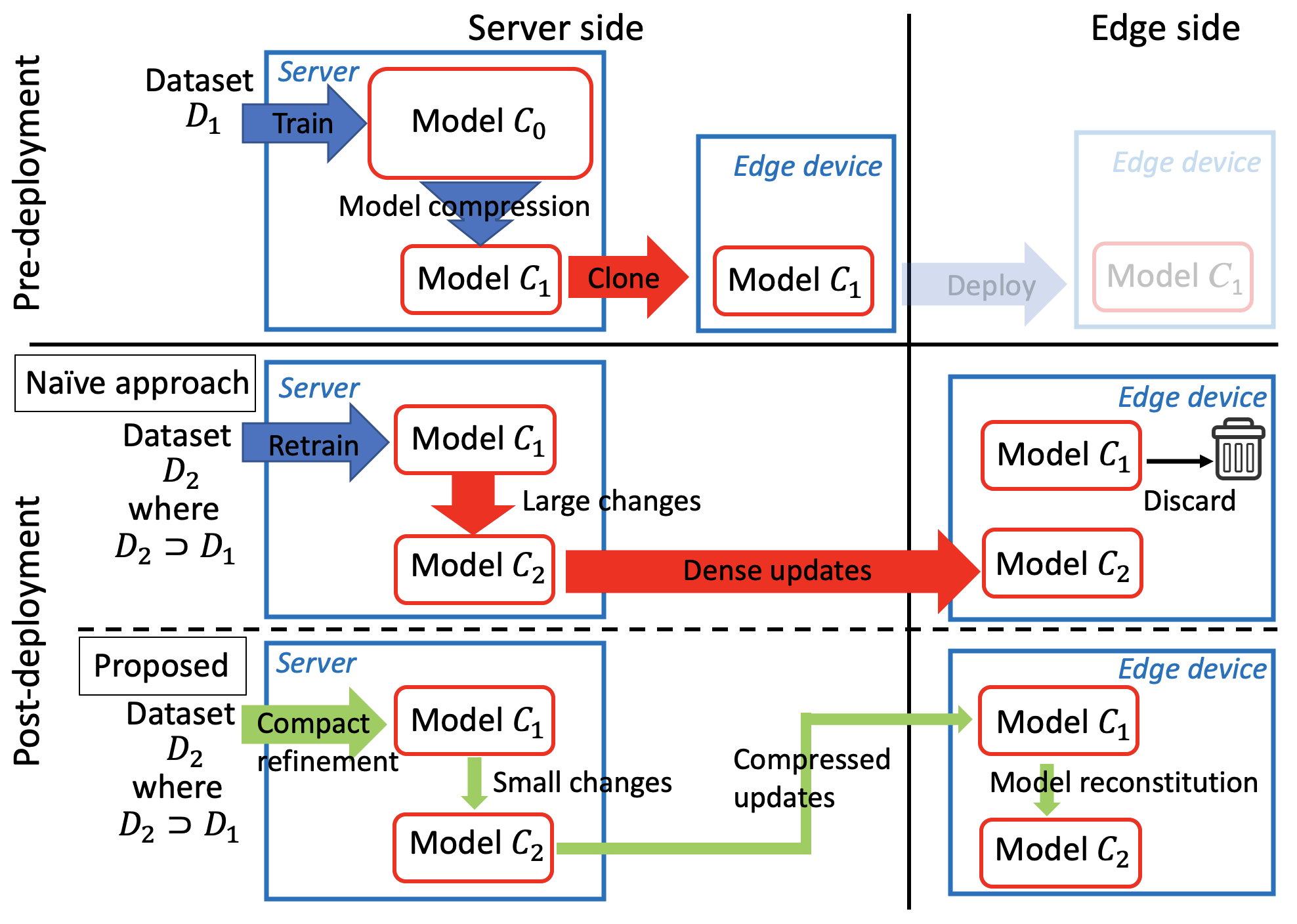}
    \caption{Illustrating update compression to alleviate bandwidth constraints for updating DNN models on the edge.}
    \label{fig:system}
\end{figure}

Another important requirement is \emph{updating} the DNN on the edge \emph{after} deployment. Numerous practical reasons, such as refining the model using new data, concept drift, or outright changes to the learning task, make updating DNN models unavoidable in machine learning applications. In this paper, we consider the specific scenario where
\begin{itemize}[leftmargin=1em,itemsep=2pt,parsep=0pt,topsep=2pt]
\item The (new) training data and supervision labels are available on the central processing system (the ``server'');
\item Communication between server and edge can be unreliable or costly, which discourages large data transfers.
\end{itemize}
Many applications fall under this setting, e.g., AI-enabled sensor networks, field robotics, autonomous driving cars, intelligent drone fleets, etc. The scenario argues for retraining the DNN on the server based on a clone of the deployed DNN, then transmitting only the new parameter values to update the model on the edge, as illustrated in Fig.~\ref{fig:system}.

\vspace{-1em}
\paragraph{Update compression}

To mitigate the communication bottleneck, it is vital to \emph{compress the model update} such that the data can be transferred as bandwidth-efficiently as possible. Note that this aim differs from compressing the model itself; indeed, even if the deployed DNN is already a compressed model, it is still worthwhile to seek to transfer as little as possible to the edge to accomplish the update, particularly when recurrent updates are required.

A simple solution is to apply lossless compression techniques~\cite{sayood2002lossless} on the model changes. However, this misses the potential gains via using more strategic model refinement that directly compresses the update during retraining.

\vspace{-1em}
\paragraph{Our contributions}


Towards the above end, we inject the goal of update compression in the model refinement step itself. We propose a \emph{compact refinement} technique based on re-parameterisation of DNN layers, which allows to express the model changes in terms of a small set of new parameters and to preserve a large part of model knowledge in the form of frozen buffers. On the edge side, a \emph{reconstitution} step integrates the learnt update package with the preserved knowledge and reconfigures the model. The proposed method is illustrated in Fig.~\ref{fig:system} and \ref{fig:pipeline}.

We examined the trade-off between update size and impact on inference accuracy to illustrate the superiority of our method over existing update compression techniques. In our classification experiment our method lifts the test accuracy of the initial model by $\approx 9\%$ with an update package equivalent to 0.5\% of the model parameters, while competing methods requires 2 to 40 times bigger update packages to achieve the same lift. Moreover, our experiments show that our model is less reliant on parameter redundancy than existing techniques, which is more suitable for compact models deployed on resource limited edge devices. 

\section{Related Work}

In this section, we briefly survey the related literature. We focus on three main research directions: \textit{Model compression} techniques that create small-scale versions of large DNNs. \textit{Update compression} methods that allow efficient transmission of model parameters in client-server architectures. \textit{Split computing} approaches that divide a DNN model into two parts that execute in the client and server cooperatively. 

\subsection{Model Compression}
\label{sec:model_compression}

The execution of deep-learning models on edge devices has attracted much attention in the last few years~\cite{chen2019deep}. Edge devices are often limited by memory, processing and power constraints that motivate the development of compressed DNN models. 

Compressing a model can benefit memory footprint and processing time, with often a minor decrease in classification performance~\cite{cheng2017survey}. There are six main categories of model compression techniques: quantisation, network slimming, weight pruning, low-rank factorisation, knowledge distillation and neural architecture search. We list below a few prominent works from each category. 

Quantisation reduces the number of bits used to represent the network weights~\cite{nagel2020up,li2017training,alizadeh2020gradient,martinez2020training,10.1007/978-3-319-46493-0_32}. Network slimming removes the least important channels from convolutional layers~\cite{tung2018clip, zhuang2018discrimination, he2017channel}. Weight pruning removes the least significant connections in the model~\cite{he2018amc}. Low-rank factorisation use matrix/tensor decomposition to estimate the informative parameters of the DNN model~\cite{kim2016compression,garipov2016ultimate,peng2018extreme,chen2018adaptive,kim2019efficient,kossaifi2019t}. Knowledge distillation trains a more compact neural network to reproduce the output of a larger network~\cite{tang2019learning,polino2018model,xu2017training,chen2016net2net,hinton2015distilling}. Neural architecture search uses a search procedure to automatically design efficient models~\cite{xie2018interleaved,sun2018igcv3,howard2019searching,ma2018shufflenet,zhang2018shufflenet,gordon2018morphnet,tan2019mnasnet,tan2019efficientnet,sandler2018mobilenetv2}. 

While the above mentioned techniques are capable of effectively compress the model size and, consequently, the update size, they are likely to be less efficient for update compression because the scenario is different: for update compression it does not require the whole model parameters to be updated which means the model can store aside useful information and only change what is necessary.

\subsection{Model Update Compression}
\label{sec:update_compression}

Besides hardware constraints, edge devices often rely on network communication to exchange data, model parameters and predictions. However, network communication can be costly, unreliable and energy-hungry. Improving communication efficiency is a research topic that has been explored in various client-server DNN architectures such as Federated Learning (FL)~\cite{mcmahan2017communication} and, more generally, distributed SGD~\cite{wang2018cooperative}. 

Our work assumes that training data is available at the server, which has the computational power to retrain the DNN model. In contrast, Federated Learning requires that training data, including class labels, be available at the client devices. These devices have the computational power to update the model locally. The server receives the model updates from multiple clients and averages these updates.

The main approaches for model update compression can be divided in three groups: low-rank update, random masking and gradient-based techniques.

The low-rank update was used as an update compression method in FL by~\cite{konevcny2016federated}. It consists in decomposing the gradient tensor in two matrices, $L$ and $R$, with a rank $r$. $L$ is generated randomly, and $R$ is optimised during training. After training, $L$ is compressed as a random seed and transmitted with $R$ to the server. 

Random masking defines a random sparsity matrix and requires the training to update the non-null entries in such a sparse matrix~\cite{konevcny2016federated}. Like the low-rank update method, a random seed defines the sparse pattern transmitted to the server with the non-zero entries.

Gradient-based techniques aim to reduce the communication burdens during training via gradient quantisation~\cite{seide20141, alistarh2017qsgd}, sparsification~\cite{konevcny2016federated, strom2015scalable}, or both~\cite{dryden2016communication, Hardy2017}. However, compressing communication during training differs from compressing communication for model update, as model training requires many iterations whereas the latter only allows for a one-off data transmission.




\subsection{Split Computation}

Split Computing (SC)~\cite{matsubara2021split} is a framework that divides the DNN model into head and tail models, which are executed in the edge device and server, respectively. SC is attractive when compressed models for edge devices cannot achieve the same level of accuracy as their full counterpart models. SC has two main limitations: the inference time becomes the sum of inference time on client and server plus the communication latency, and the need for a ``bottleneck'' layer in the first model layers. The bottleneck layer provides a cutpoint for the head and tail models with a compact representation that reduces communication overhead. 

The straightforward splitting of the DNN as suggested by~\cite{lane2016deepx,kang2017neurosurgeon,jeong2018computation} can result in either transferring a large part of the processing burden to the edge device or transmitting a larger volume of data on the network. Distilling the head section of DNN and introducing a bottleneck within the distilled head model as suggested in~\cite{matsubara2019distilled} can mitigate this problem and decrease the computational cost and the required bandwidth considerably.

\section{Problem setting}

We consider the problem of performing bandwidth efficient updating for a DNN on an edge device deployed in the field. As depicted in Fig.~\ref{fig:system}, the setting consists of a central server and an edge device, between which there is a communication bandwidth limit. We assume that the central server has no restriction in computational resources while the edge device has limited resources and can only perform inference and light-duty computation.

\vspace{-1em}
\paragraph{Pre-deployment}

For concreteness, we focus on classifier models $C(\cdot|\theta)$ with parameters $\theta$. Also, since our work concentrates on update compression and not model compression (see Fig.~\ref{fig:system} on the distinction), we assume that $C(\cdot|\theta)$ is already a lightweight model suitable for edge devices.

Let $D_1 = \{(\bm{X}_i, Y_i)\}_{i=1}^{N_1}$ denote the initial training dataset. At the pre-deployment stage, training $C$ with $D_1$ results in model parameters $\theta_1$ 
\begin{equation}
    \theta_1 = \argmin_{\theta} \ell(\theta, D_1), 
\end{equation}
where $\ell$ is some loss function for training $C$, such as cross-entropy loss. The classifier $C(\cdot | \theta_1)$ is then loaded onto the edge device and deployed to the field.

\vspace{-1em}
\paragraph{Post-deployment}

After deployment, the central server has accumulated a larger dataset $D_2 = \{(\bm{X}_i, Y_i)\}_{i=1}^{N_2}$, where $N_2 > N_1$ and $D_2 \supset D_1$. The goal is to update $C$ on the edge device, currently with parameters $\theta_1$, with the new dataset $D_2$, subject to a data transmission bottleneck.

Let $\Delta$ denote the update package to be sent from the server to the edge device. A basic method is to retrain $C$ using $D_2$, and let $\Delta$ be the set of all model parameters, \ie,
\begin{equation*}
    \Delta = \argmin_{\theta} \ell(\theta, D_2).
\end{equation*}
At the edge upon receiving $\Delta$, $\theta_1$ is discarded and we simply equate $\theta_2 = \Delta$ and insert it into $C(\cdot | \theta_2)$ to complete the model update. However, in this case $\Delta$ will always be the set of all parameters even for a small incremental update. The key to generating more compact updates lies in reusing the learnt knowledge embedded in $\theta_1$, as we will show next.

\section{Update compression}\label{sec:methods}

\begin{figure*}[t]
    \centering
    \includegraphics[width=0.8\textwidth]{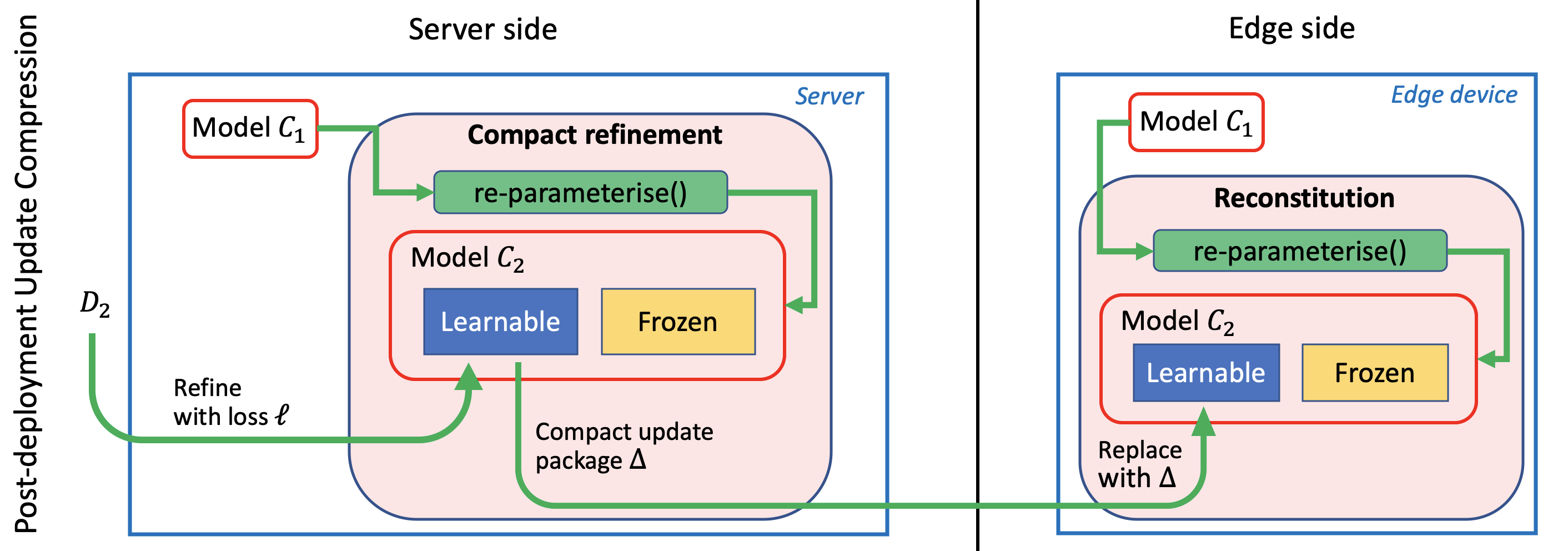}
    \caption{A schematic pipeline of the proposed post-deployment update compression framework. To achieve compactness of the update package, we propose to re-parameterise the model into a learnable part and a frozen part. The learnable parameters facilitates the compact refinement and the frozen data enables the edge device to recycle knowledge from existing model.}
    \label{fig:pipeline}
\end{figure*}

Instead of densely retraining the whole model, we would like a \emph{compact refinement} algorithm that produces a compact update package $\Delta$ using the current model $C(\cdot|\theta_1)$ and new dataset $D_2$
\begin{equation}
    \Delta = \text{compact-refine}(C(\cdot | \theta_1), D_2), 
\end{equation}
so that it can be efficiently transmitted to the edge device. At the edge side, upon receiving the compact update package $\Delta$, the current model $C(\cdot|\theta_1)$ incorporates $\Delta$ and reconstitutes the updated model 
\begin{equation}
    C(\cdot|\theta_2) = \text{reconstitute}(C(\cdot | \theta_1),  \Delta).
\end{equation}
The overall pipeline of such a update compression scheme is demonstrated in Fig.~\ref{fig:pipeline}.

\subsection{Compact refinement}
To achieve compact refinement, we first re-parameterise the current network in a way such that 
\begin{enumerate}
    \item the new learnable parameters are much fewer than the original ones; and
    \item the model can be effectively updated by refining the new parameters with the new dataset $D_2$.
\end{enumerate}

To this end, we focus on the network's convolutional (Conv) layers and fully connected (FC) layers, as mainstream DNNs largely consist of these two types of layers. Other types of layers, such as batch normalisation~\cite{ioffe2015batch}, are not targeted for update compression in our method.


\paragraph{Low Rank Approximation (LRA)} We first describe the compact refinement algorithm using a baseline re-parameterisation method LRA before we introduce the proposed techniques. Let $\phi_l \subset \theta$ denote the weight tensor of layer $l$, which is either a Conv or FC layer. Note that $\phi_l$ does not include the bias terms $\bm b_l$ of the layer. If the layer has bias terms, they are not aimed for update compression in our method and remain as learnable parameters. We firstly re-arrange $\phi_l$ to a matrix of size $o \times i$. For Conv layers, $o$ is the number of output channels and $i$ is the product of the number of input channels and the size of the kernel along each dimension, \eg, the kernel height and the kernel width for the 2D case. For FC layers, $o$ is simply the number of output features and $i$ the input features. 

The matrix $\phi_l$ is then decomposed using SVD:
\begin{equation}\label{eq:svd}
    \phi_l = U \cdot \text{diag}(\bm s) \cdot V^T, 
\end{equation}
where $U \in \mathbb{R}^{o \times m}$, $V \in \mathbb R^{i \times m}$, $m = \min(o, i)$, $\bm s \in \mathbb R^{m \times 1}$ is the vector of singular values sorted in descending order, and $\text{diag}(\bm s)$ is the $m \times m$ square matrix of $0$s except for the diagonal entries which are specified by $\bm s$. 

For either Conv or FC layers, the arrangement of $\phi_l$ can be viewed as $o$ rows of filters with length $i$. Consequently, the decomposed $U$, $V$ and $\bm s$ can be viewed as a mapping matrix, a matrix of base filters, and a weight vector respectively, the latter indicating the importance of each base filter.

The decomposition of $\phi_l$ can result in even more parameters if $U, \bm s$, and $V$ are directly used as new parameters. To achieve compactness, we adopt LRA to encode the original parameters. Let $U_{1:r}$ denote the $o \times r$ matrix which consists of the first $r$ columns of $U$, and $\bm s_{1:r}$ denote the $r \times 1$ vector of the first $r$ entries in $\bm s$.
We then define a new set of learnable parameters for the layer
\begin{equation}
    \phi_l' = \{L, R\},
\end{equation}
where
\begin{equation}\label{eq:LR}
    L =  U_{1:r} \cdot \text{diag}(\bm s_{1:r}), \quad 
    R = V_{1:r}^T. 
\end{equation}

\begin{figure}
    \centering
    \includegraphics[width=\linewidth]{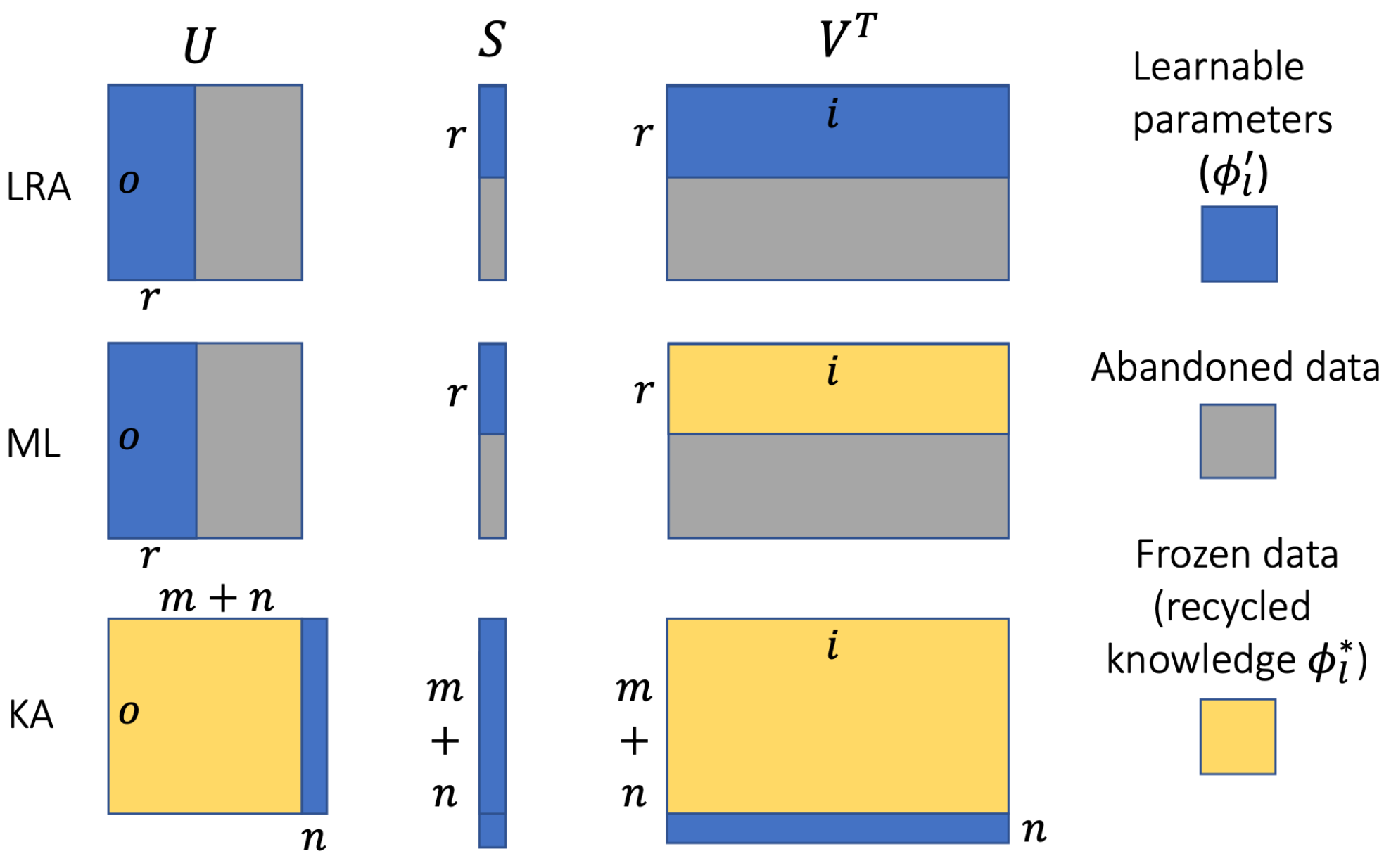}
    \caption{Illustration of different re-parameterisation methods.}
    \label{fig:methods}
\end{figure}

Note that this does not change the network's architecture. After a Conv or FC layer is re-parameterised, for each forward propagation during the refinement process, the layer's weight tensor $\phi_l$ is firstly computed with $\phi_l'$ using a recover-weight() routine
\begin{align}
    \phi_l &\leftarrow \text{recover-weight}(\phi_l') \\
    &= L \cdot R.
\end{align}
The layer then performs normal forward processing. For backward propagation in such layers, only gradients with respect to $\phi_l'$ are computed instead of $\phi_l$, since $\phi_l$ is now an intermediate variable rather than a leaf variable in the computational graph. The LRA re-parameterisation method is conceptually illustrated in Fig.~\ref{fig:methods} row 1. 

By re-parameterising all Conv and FC layers, the network's new learnable parameters become
\begin{equation}
    \theta' = \left(\theta \setminus \{\phi_l\}_{l=1}^{\cal L} \right) \cup \{\phi_l'\}_{l=1}^{\cal L},
\end{equation}
where $\cal L$ is the total number of Conv and FC layers in the network. Note that $\theta'$ includes the network's original parameters that are not targeted for compact refinement, \eg, bias terms, batch-norm parameters. 

Finally, the update package $\Delta$ is obtained via refining $\theta'$ from its current values with $D_2$
\begin{equation}
    \Delta = \argmin_{\theta'} \ell(\theta', D_2).
\end{equation}

The LRA approach could use a small rank $r$ to achieve compactness
\begin{equation}
    |\Delta| = |\theta'| \ll |\theta|.
\end{equation}
However, since it does not reuse any existing knowledge from $\theta_1$ it can result in severe underfitting if $r$ is too small. LRA can be viewed as a network compression method which is similar to previous works~\cite{jaderberg2014speeding, lebedev2015speeding, denton2014exploiting, kim2016compression, lin2018holistic}. It is thus considered as a baseline method in our experiments. 

\paragraph{Mapping Learning (ML)}
To further reduce the number of learnable parameters used in refinement while performing effective update, we propose to recycle part of previously learnt knowledge embedded in $\theta_1$. We obtain $L$ and $R$ the same way as Eq.~\eqref{eq:LR} with rank $r$. The matrix $R$ consists of the most important $r$ rows of base filters learnt from $D_1$. It is thus possible to utilise these base filters and only learn a new mapping for the  model to adapt to dataset $D_2$. 

Therefore, the re-parameterisation of $\phi_l$ should produce a set of learnable parameters $\phi_l'$ and a set of frozen data $\phi_l^*$ which stores the recycled knowledge:
\begin{equation}
    \phi_l' = L, \quad \phi_l^* = R.
\end{equation}

The frozen data $\phi_l^*$ is stored in the network as a buffer, which participates in forward computation but does not require gradient computation in back-propagation. 
Consequently, the recover-weight() routine is given by
\begin{equation}\label{eq:ml_update}
    \text{recover-weight}(\phi_l', \phi_l^*) = \phi_l' \cdot \phi_l^* = L \cdot R.
\end{equation}

\paragraph{Knowledge Augmentation (KA)}
To maximally preserve previously learnt knowledge in order to minimise the update size, we propose to freeze both $U$ and $V$ in Eq.~\eqref{eq:svd}, and only refine on $\bm s$ as free parameters. However, the base filters $V$ and its mappings $U$ learnt from $D_1$ are likely to generalise poorer than they could have if they were learnt from $D_2$, because a larger training set is usually beneficial as it is less likely to cause overfitting. 


To let the model learn to adapt to $D_2$ whilst preserving current knowledge $U$ and $V$, we propose to augment $U$ and $V$ with additional columns for capturing the ``knowledge drift" between $D_1$ and $D_2$. Let $U' \in \mathbb R^{o \times n}$ denote the augmenting vectors and $[U,U'] \in \mathbb R^{o \times (m+n)}$ denote the augmented mapping matrix.
Similarly let $V' \in \mathbb R^{i \times n}$ be the augmenting vectors and $[V,V'] \in \mathbb R^{i \times (m+n)}$ be the augmented filter matrix. Note that both $U$ and $V$ are frozen and only $U'$ and $V'$ are learnable parameters. 

Let $\bm s' \in \mathbb R^{(m+n)\times 1}$ denote the augmented weight vector where the whole vector is treated as learnable parameters.
The layer is thus re-parameterised as 
\begin{align}
    \phi_l' &= \{U', \bm s', V'\}, \quad \phi_l^* = \{U, V\},
\end{align}
and the recover-weight() routine is 
\begin{equation}
    \text{recover-weight}(\phi_l', \phi_l^*) = [U, U'] \cdot \text{diag}(\bm s') \cdot [V,V']^T. 
\end{equation}
An illustration of this re-parameterisation method is given in Fig.~\ref{fig:methods} row 3. The motivation of this design is to maximally reuse previous knowledge by freezing $U$ and $V$, and at the same time allowing them to be updated by learning the augmenting vectors and incorporating them using the re-learned weights $\bm s'$. 

After re-parameterisation we initialise $\phi_l'$ in a way such that the refinement process starts from the current accuracy level. The refinement process starts with a weight tensor 
\begin{equation}
    \phi_l \leftarrow [U, U'] \cdot \text{diag}(\bm s') \cdot [V,V']^T
\end{equation}
which could be abruptly different from the original $\phi_l$ because $U', \bm s'$ and $V'$ are initialised randomly. To avoid a significant drop in accuracy at the beginning of refinement, we let $\bm s_{1:m}' = \bm s$ and $U', V'$ and $\bm s_{m+1:m+n}'$ be small random values that are close to 0 for initialisation. 

\begin{algorithm}[h]
\caption{compact-refine$(C(\cdot|\theta), D)$}
\label{alg:compact}
\begin{algorithmic}[1]
\STATE $\theta' \leftarrow \theta$
\FOR{$l$ in $\{1,...,\cal L\}$ }
\STATE $\phi_l \leftarrow$ layer $l$ weight tensor in $\theta_1$
\STATE $\phi_l', \phi_l^* \leftarrow \text{re-parameterise}(\phi_l)$ 
\STATE $\theta' \leftarrow (\theta' \setminus \phi_l) \cup \phi_l'$
\ENDFOR
\STATE $\Delta \leftarrow \argmin_{\theta'} \ell(\theta', D)$ \quad 
\RETURN $\Delta$
\end{algorithmic}
\end{algorithm}

The overall procedure of compact refinement is summarised in Algorithm~\ref{alg:compact}, with the re-parameterise() routine and the recover-weight() routine depending on the specific method employed. 

\subsection{Model reconstitution}

Once the edge device receives $\Delta$, it then incorporates $\Delta$ to update its current model. This process is described in Algorithm~\ref{alg:reconstitute}. For all non Conv or FC layers, the edge device updates their parameters with new values from $\Delta$. For Conv and FC layers, their weights are recovered using the new values of the learnable parameters $\phi_l'$ from $\Delta$ and the values of the frozen data $\phi_l^*$ obtained by decomposing the current model weights $\theta_1$. 

\begin{algorithm}[h]
\caption{reconstitute$(C(\cdot|\theta_1), \Delta)$}
\label{alg:reconstitute}
\begin{algorithmic}[1]
\STATE $\theta_2 \leftarrow \{\text{all non Conv/FC layers in }\Delta\}$
\FOR{$l$ in $\{1,...,\cal L\}$}
\STATE $\phi_l \leftarrow$ layer $l$ weight tensor in $\theta_1$
\STATE $\phi_l', \phi_l^* \leftarrow \text{re-parameterise}(\phi_l)$
\STATE $\phi_l^{\Delta} \leftarrow $ layer $l$ free parameters in $\Delta$
\STATE $\phi_l \leftarrow$ recover-weight$(\phi_l^{\Delta}, \phi_l^*)$
\STATE $\bm b_l \leftarrow $ layer $l$ bias in $\Delta$
\STATE $\theta_2 \leftarrow \theta_2 \cup \{\phi_l, \bm b_l \}$
\ENDFOR
\RETURN $C(\cdot|\theta_2)$
\end{algorithmic}
\end{algorithm}

An additional desirable feature of the proposed KA method is that, once the edge device reconstitutes the updated model the augmenting vectors are integrated into the weight tensor by the recover-weight() routine and thus can be discarded. This prevents the edge device to store more and more augmenting vectors as a result of repetitive updates, which eliminates the requirement of maintenance over time for expanding its memory capacity.





\section{Experiments}

In this section we conduct various experiments to evaluate the performance of update compression with the proposed methods, as well as competing methods.

\subsection{Competitor methods}

We compared the proposed update compression algorithms to two techniques used in Federated Learning, Random Mask (RM)~\cite{konevcny2016federated} and Low-Rank Update (LRU)~\cite{konevcny2016federated}, as well as a compression method Filter Pruning (FP)~\cite{li2017pruning}.

\paragraph{RM} For the RM method we generate a random binary mask for the entire model parameters. The proportion of parameters that are selected by the mask is controlled by a hyper-parameter $P$. We then refine the model on $D_2$ by updating only the selected parameters while keeping the unselected ones frozen. The update package size is counted as the number of selected parameters. We ignore the size of the mask as it can be represented by a random seed. 

\paragraph{LRU} To implement LRU we firstly decompose the weight tensor of each Conv and FC layer to two matrices $L$ and $R$ with rank $r$ in the same way as Eq.~\eqref{eq:LR}. We then replace the values of $R$ with random numbers and fix them. During refinement $L$ is initialised with small values and $L \cdot R$ is added to the layer's original weight tensor, \ie, $L \cdot R$ learns the changes of the current weight tensor, as described in~\cite{konevcny2016federated} Sec.2. The update size of LRU is all the $L$ matrices and other free parameters (such as bias, Batch-Norm layers, etc). Again we ignore the size of the random matrices $R$ as they can be represented by random seeds. 

\paragraph{FP} The FP method prunes whole filters as well as their connected feature maps in the network. The size of the pruned model is dependent on a compression rate $c$ and the layers selected for the pruning plan. In this experiment we only consider Conv and batch normalisation layers in the pruning plan. We count all remaining parameters after pruning into the update size. 




\subsection{Update compression for classification models}

\begin{figure*}[h]
    \centering
    \begin{subfigure}[b]{0.45\textwidth}
         \centering
         \includegraphics[width=\linewidth]{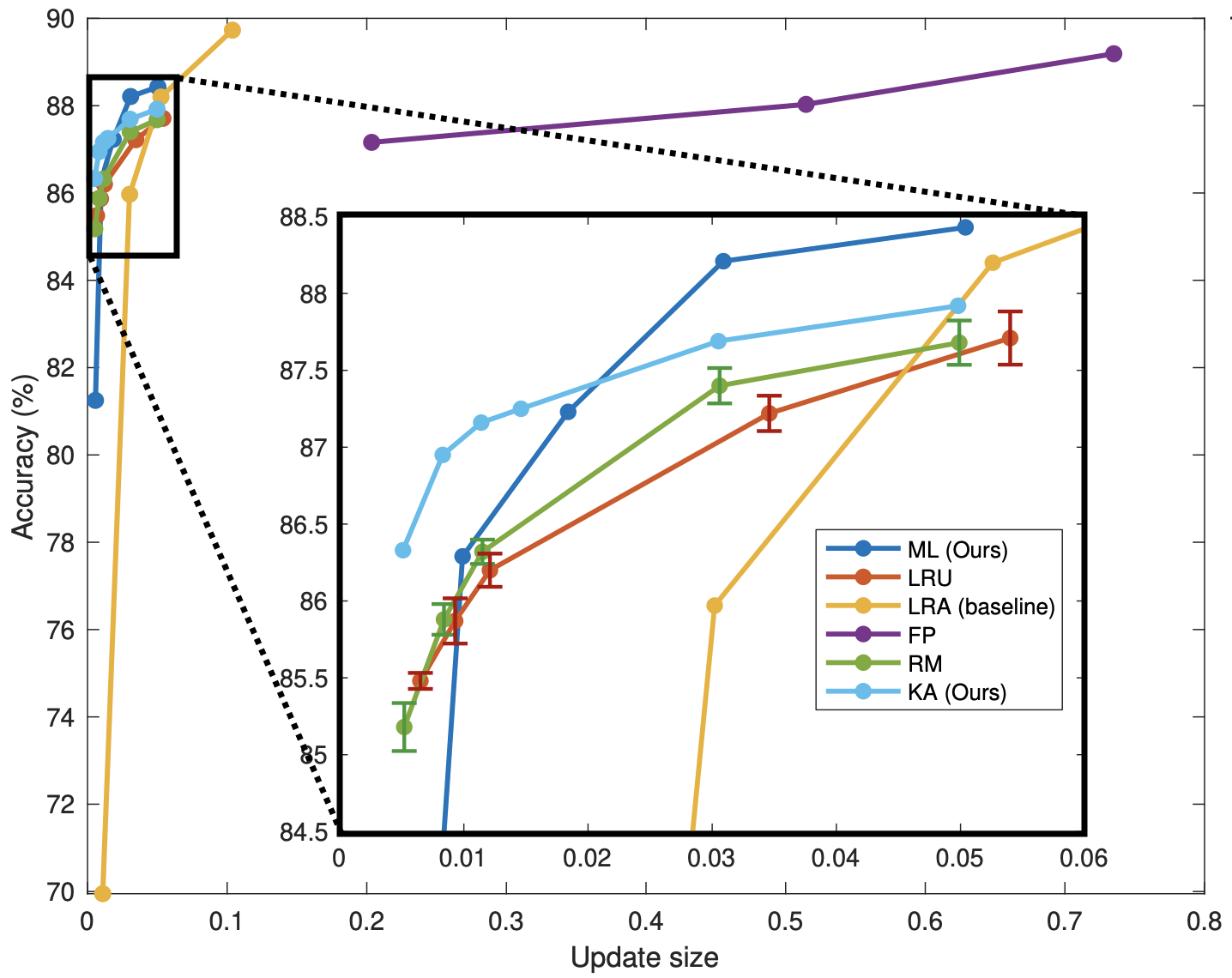}
         \caption{CIFAR10 classification with ResNet18. Results of RM and LRU are averages of 6 runs and their 90\% confidence intervals are plotted.}
        \label{fig:resnet18}
     \end{subfigure}
     \hfill
     \begin{subfigure}[b]{0.44\textwidth}
         \centering
         \includegraphics[width=\linewidth]{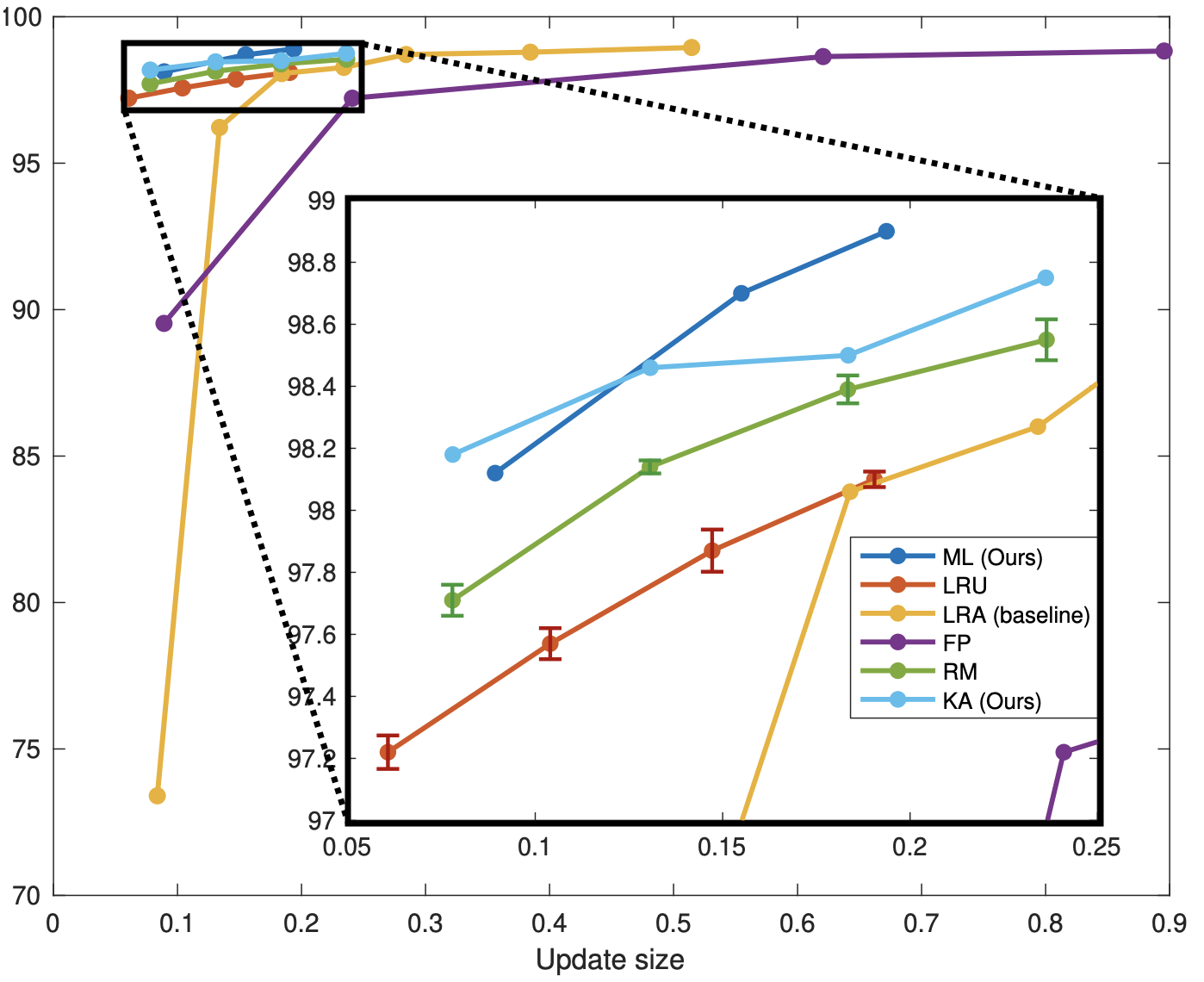}
         \caption{MNIST classification with tiny-ResNet. Results of RM and LRU are averages of 6 runs and their 90\% confidence intervals are plotted.}
        \label{fig:resnet-tiny}
     \end{subfigure}
     \caption{The update size versus test accuracy after refinement. Update size is measured as a proportion between the number of new parameters ($|\theta'|$) and the number of original parameters ($|\theta|$).}
     \label{fig:classification}
\end{figure*}

We evaluate update compression methods for two classification tasks. The first one uses the CIFAR10~\cite{krizhevsky2009learning, cifar10} dataset on a ResNet18~\cite{he2016deep} model and the second one classifies the MNIST~\cite{lecun1998gradient,mnist} dataset with a modified tiny-ResNet model. Since we are mainly concerned with the power- and bandwidth-constrained application, the CIFAR10 and MNIST datasets with low-resolution images are a good choice The tiny-Resnet architecture is composed of a feature extractor with three layers, each layer consists of a basic residual block and a classifier that include one fully connected layer. Similarly, the VGG-tiny architecture is comprised of a feature extractor that contains a sequence of convolutional layers with 16, 16, 32, and 64 outputs, each is followed by a max-pooling layer. Besides, one fully connected layer is on top of them. For both tasks we use a proportion $p$ of the training set as $D_1$ and the whole training set as $D_2$. The model is initially trained with $D_1$ and then updated using $D_2$ with various update methods. 
We then test the updated model with the testing set and report its accuracy against the update size. The proportion for $D_1$ is $p=0.2$ for the CIFAR10 experiment and $p=0.02$ for MNIST, as MNIST is a much easier dataset and a higher $p$ would make the accuracy gap between the initial model and the refined model too small for the experiment. The initial test accuracy of the model trained with $D_1$ is 77.5\% for ResNet18 on CIFAR10 and 95.5\% for tiny-ResNet on MNIST. 
The results of this experiment is shown in Fig.~\ref{fig:classification}. Note that in this experiment the purpose is not to push the test accuracy as high as possible, but to compare the updating efficiency among different methods, \ie, achieving high accuracy with a small update size. 

Fig.~\ref{fig:resnet18} shows that KA has the highest accuracy for update size below 2\%, while ML performs best with update size between 2\% to 5\%. The curves of RM and LRU are much more to the right hand side than KA's, which means in order to achieve the same accuracy, RM and LRU requires 2 to 3 times larger update packages than KA. 
The low rank approximation nature of ML makes it deteriorates quickly when update size shrinks below 1\%. On the other hand, KA is much more robust for small update size thanks to its augmentation mechanism. For the compression method FP, its accuracy at update size 20\% is just similar to that of KA at update size 1\%. Results in Fig.~\ref{fig:resnet-tiny} tell a similar story where KA has the highest accuracy for the smallest update, closely followed by ML. In contrast, it takes RM and LRU at least twice as big an update size as KA's for the same accuracy. 

Note that in both experiments the LRA baseline is capable to achieve the highest accuracy when given enough parameter size but performs poorly at a low update size. This indicates that knowledge recycling has its pros and cons: while it can be super efficient in terms of update size, methods like LRA without preserving any old knowledge can achieve higher accuracy when given enough parameters.  

\begin{table*}[t] 
\begin{center} \small
\begin{tabular}{llrrrrrr} \hline

\hline
\textbf{Model} & \textbf{Method} & \textbf{Update size} & \textbf{Update \%} & \textbf{Model size} & \textbf{Acc. lift} & \textbf{Refined acc.} & \textbf{Init acc.}  \\
\hline
VGG-small & KA & 7413 & 5.44 & 136266 & \textbf{2.28} & 78.68 & 76.4 \\ 
VGG-small & RM & 7413 & 5.44 & 136266 & 0.02 & 76.42 & 76.4 \\ 
VGG-small & LRU & 7878 & 5.78 & 136266 & 1.45 & 77.85 & 76.4 \\ 
\hline
VGG-medium & KA & 14672 & 2.72 & 539786 & \textbf{2.83} & 80.97 & 78.14 \\ 
VGG-medium & RM & 14681 & 2.72 & 539786 & 0.07 & 78.21 & 78.14 \\ 
VGG-medium & LRU & 15638 & 2.90 & 539786 & 1.44 & 79.58 & 78.14 \\ 
\hline
MobileNetV2 & KA & 122632 & 5.34 & 2296922 & \textbf{3.67} & 87.92 & 84.25 \\ 
MobileNetV2 & RM & 122660 & 5.34 & 2296922 & 3.18 & 87.43 & 84.25 \\ 
MobileNetV2 & LRU & 128734 & 5.60 & 2296922 & 3.54 & 87.79 & 84.25 \\ 
\hline
ResNet18 & KA & 127939 & 1.14 & 11173962 & \textbf{9.67} & 87.16 & 77.49 \\ 
ResNet18 & RM & 128378 & 1.15 & 11173962 & 8.97 & 86.46 & 77.49 \\ 
ResNet18 & LRU & 135670 & 1.21 & 11173962 & 9.02 & 86.51 & 77.49 \\ 
\hline
ResNet50 & KA & 310656 & 1.32 & 23520842 & \textbf{6.57} & 84.34 & 77.77 \\ 
ResNet50 & RM & 312837 & 1.33 & 23520842 & 4.61 & 82.38 & 77.77 \\ 
ResNet50 & LRU & 327185 & 1.39 & 23520842 & 6.16 & 83.93 & 77.77 \\ \hline

\hline
\end{tabular}
\end{center}
\caption{Accuracy lift on different models. Update size and model size are both measured in number of parameters.}
\label{tab:model_v_lift}
\end{table*}

\subsection{The effect of parameter redundancy}

Besides knowledge recycling, we suspect that another factor contributing to the effectiveness of update compression is parameter redundancy, which allows the model to adapt to the new dataset by updating only a small fraction of parameters. However, for edge devices models deployed on them are usually optimised, distilled, or compressed to perform the target task well enough while minimising resource consumption. Therefore, update compression methods should be robust to the lack of parameter redundancy. 

\begin{figure}[h]
    \centering
    \includegraphics[width=0.8\linewidth]{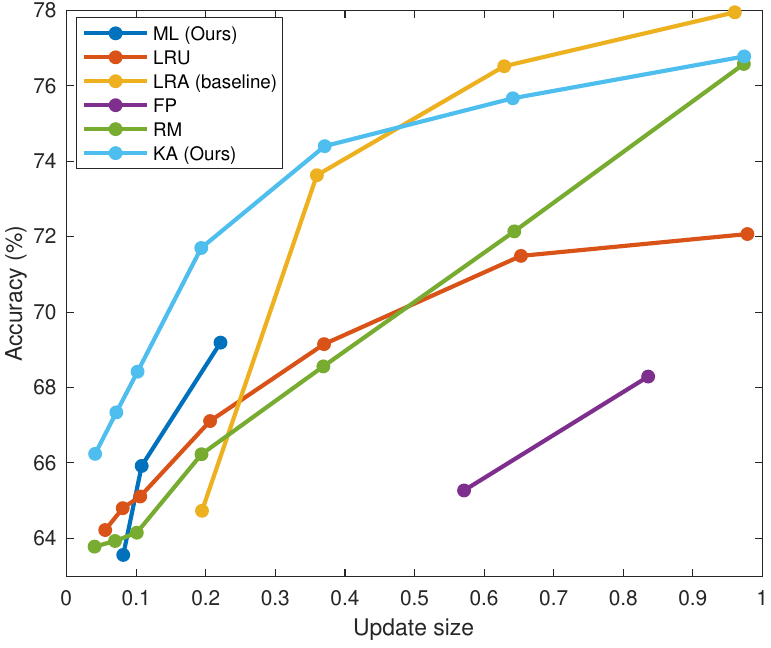}
    \caption{The update size versus test accuracy after refinement on VGG-tiny. }
    \label{fig:VGG-tiny}
\end{figure}

We repeat the CIFAR10 experiment on a customised VGG-tiny~\cite{simonyan2015very} model, a much smaller model than ResNet18 with only 26554 parameters, to test different updating methods on compact models that do not have a lot of parameter redundancy. The result in Fig.~\ref{fig:VGG-tiny} shows that the gap between KA and the two competing methods RM and LRU is clearly wider than that in ResNet18. This suggests that with a small model the lack of parameter redundancy has restricted the performance of RM and LRU, while the proposed KA is less affected by the small model size, which is attractive for resource limited edge devices. 

To further test this hypothesis, we compare KA, RM and LRU on different models with parameters ranging from 136K to 23M. For each model, we set the hyper-parameter $n$ of KA to 3 and let the update size of RM and LRU to be similar to but no less than KA's. We use 20\% of CIFAR10 training set as $D_1$ and 100\% as $D_2$. We report the accuracy lift as well as other experimental details in Table~\ref{tab:model_v_lift}.

For all models in Table~\ref{tab:model_v_lift} regardless of their size, KA outperforms RM and LRU consistently with an equal or smaller update size. Furthermore, as the model becomes more and more compact the accuracy lifts of all methods tend to be smaller as well, supporting our hypothesis about parameter redundancy. This effect is especially obvious for RM, which can hardly lift accuracy for small models. In contrast, KA is less reliant on parameter redundancy and remains effective for models of all sizes. 

\subsection{Effectiveness of knowledge renewal}

All three methods of KA, RM, and LRU reuses significant old knowledge to achieve update compression. 
RM keeps all unselected parameters and only learns selected ones to update the model, while LRU renews the old model by learning to approximate the residuals of the old parameters. We evaluate the effectiveness of knowledge renewal amongst the three different techniques in the circumstance that the model is compact and there is little parameter redundancy. We train a VGG-tiny model with different initial training set size, \ie, $|D_1|$ ranges from 0.1 to 0.7 of $|D_2|$. We then perform update compression to the model with different techniques and report their accuracy lifts after refinement with $D_2$. For KA the hyper-parameter $n$ is set to 1. For RM and LRU their update size are set to be as close as but not less than KA's.

\begin{figure}[h]
    \centering
    \includegraphics[width=0.8\linewidth]{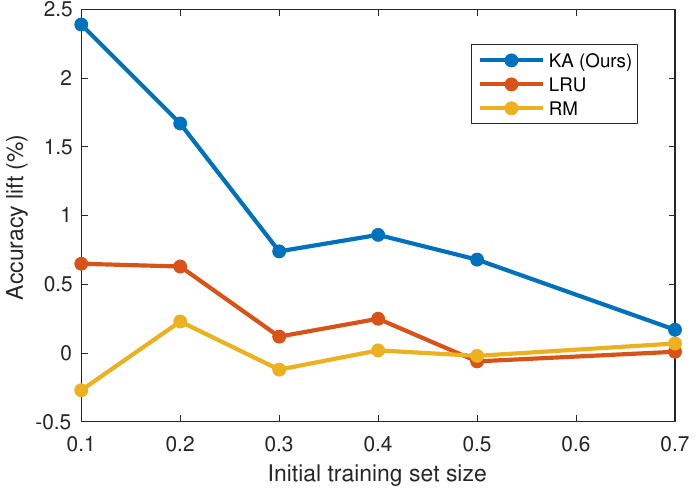}
    \caption{Initial training set size vs accuracy lift on VGG-tiny. The initial training set size is computed by $|D_1|/|D_2|$.}
    \label{fig:iniset_v_lift}
\end{figure}

Fig.~\ref{fig:iniset_v_lift} shows that while the gap between the initial set $D_1$ and the target set $D_2$ enlarges, KA is more effective in learning new knowledge to update the model than RM and LRU, even when parameter redundancy is lacking. This implies the KA update strategy would benefit applications where $D_2$ grows ever larger over time.

\section{Limitations}

The proposed update compression methods only targets Conv and FC layers. It is unable to handle models that largely consist of other types of layers, such as RNN~\cite{elman1990finding}, LSTM~\cite{hochreiter1997long} and Transformer~\cite{vaswani2017attention}.
Due to limited resources and paper space, we have also limited our experiments to classification tasks and small to medium size models, the effect of practical application of our methods outside of these settings requires further testings. 

\vspace{-0.2em}
\section{Conclusion}

In this paper we address the problem of server-to-edge device model update with a limited communication bandwidth. The proposed update compression framework achieves superior communication efficiency via novel compact refinement and reconstitution techniques. Experiments further show that our method is robust to the lack of parameter redundancy.

\section*{Acknowledgements}

We gratefully acknowledge generous funding from SmartSat CRC, Research Program 2: Advanced Satellite Systems, Sensors and Intelligence.

Tat-Jun Chin is SmartSat CRC Profesorial Chair of Sentient Satellites.

\vfill

\pagebreak

{\small
\bibliographystyle{ieee_fullname}
\bibliography{egbib}
}

\end{document}